\documentclass[aoas,preprint]{imsart}

\RequirePackage{amsthm,amsmath,amsfonts,amssymb}
\RequirePackage[authoryear]{natbib}
\RequirePackage{graphicx}
\RequirePackage{rotating,booktabs}
\RequirePackage{longtable}
\RequirePackage{multirow}
\RequirePackage{hhline}
\RequirePackage{microtype}
\RequirePackage{array}
\newcolumntype{M}[1]{>{\centering\arraybackslash}m{#1}}
\startlocaldefs
\newtheorem{theorem}{Theorem}
\newtheorem{lemma}[theorem]{Lemma}
\graphicspath{{fig/}}

\endlocaldefs

\begin{document}
	
	\begin{frontmatter}
		\title{Regularized Text Logistic Regression: Key Word Detection and Sentiment Classiﬁcation for Online Reviews}
		\runtitle{Regularized Text Logistic Regression}
		
		\begin{aug}
			\author[A,B]{\fnms{Ying} \snm{Chen}\ead[label={e1}]{matcheny@nus.edu.sg}},
			\author[C]{\fnms{Peng} \snm{Liu}\ead[label=e2]{liu.peng@u.nus.edu}}
			\and
			\author[D]{\fnms{Chung Piaw} \snm{Teo}\ead[label=e3]{bizteocp@nus.edu.sg}}
			\address[A]{Department of Mathematics, National University of Singapore, \printead{e1}}
			
			\address[B]{Risk Management Institute, National University of Singapore, \printead{e1}}
			
			\address[C]{Department of Statistics \& Applied Probability, National University of Singapore, \printead{e2}}
			
			\address[D]{Institute of Operations Research and Analytics, National University of Singapore, \printead{e3}}
			
		\end{aug}
		
		\begin{abstract}
			Online customer reviews have become important for managers and executives in the hospitality and catering industry who wish to obtain a comprehensive understanding of their customers' demands and expectations. We propose a Regularized Text Logistic (RTL) regression model to perform text analytics and sentiment classification on unstructured text data, which automatically identifies a set of statistically significant and operationally insightful word features, and achieves satisfactory predictive classification accuracy. We apply the RTL model to two online review datasets, Restaurant and Hotel, from TripAdvisor. Our results demonstrate satisfactory classification performance compared with alternative classifiers with a highest true positive rate of 94.9\%. Moreover, RTL identifies a small set of word features, corresponding to 3\% for Restaurant and 20\% for Hotel, which boosts working efficiency by allowing managers to drill down into a much smaller set of important customer reviews. We also develop the consistency, sparsity and oracle property of the estimator. 
			
		\end{abstract}
		
		
	\end{frontmatter}
	

	\section{Introduction}\label{sec:intro}
	
	Online customer reviews have never before been so important in the hospitality and catering industry. With the wide availability of data on the Internet, potential customers are able to access opinions about a particular destination, product, or service from people around the world via online review platforms such as TripAdvisor.  By sharing and exchanging experiences with each other via a word-of-mouth (WOM) communication channel, customers can benefit from these online social ties and are able to make more informed decisions, especially in the field of tourism, see \cite{Abbasi2008}. From the operating and managerial standpoint of hotels and restaurants, the focus is usually on the polarity of customer reviews, and also on the key drivers behind the customer's positive or negative sentiment. Managing and understanding the online reviews for restaurants and hotels has become essential to pave the path to the greater, long term success of a business. By viewing the reviews, managers can glean insights into their branding, positioning, and quality improvement, which has a direct influence on many aspects of the business, such as hotel occupancy rate, see \cite{WOM2016}. 
	
	Nevertheless,  to discover and quantify valuable pieces of information, in particular the important word features, from large amounts of unstructured text is a nontrivial task. Even though management can gain a competitive edge by reviewing each and every customer's feedback, manual exhaustive text perusal is time consuming and inefficient, even for a moderately sized corpus. Meanwhile,  only reading a few randomly selected reviews provides an incomplete understanding, while crowdsourcing by human agents is prone to subjective bias in judgment. Even if the task of feature selection could be completed via the use of certain statistics, such as word frequency (i.e., focusing on more frequent words), the selected word features may not necessarily have statistical association with review sentiment. Consequently, we provide a statistical model to  automatically detect a set of statistically significant and operationally meaningful word features that can help managers and executives to improve working efficiency and pinpoint the key drivers of customers' polarity. 
	
	Sentiment classification aims to label reviews with positive or negative tone, where automated text analytics is used as a systematic and cost effective tool. The classification is often either based on lexicon or a supervised learning approach, see \cite{ATA_2018}. Lexicon is a validated database or dictionary containing a list of words annotated with corresponding semantic orientation (i.e., polarity and/or strength), see \cite{Taboada2011}. Meanwhile, a review is classified according to the frequency of positive and negative words in the corpus. Although lexicon-based classification has been widely used thanks to its intuitive nature, there are some limitations when the lexicon is developed outside its domain. Meanwhile, supervised learning approaches train classifiers to predict the customer's polarity by learning from labelled instances, and then capturing the underlying pattern and dependence in the reviews. 
	
	The popular machine learning classifiers include logistic regression, Support Vector Machine (SVM) and Naive Bayes (NB), see \cite{Twitter_2016}. Among others, logistic regression is often used as a baseline model in sentiment classification because of its ease of interpretation. Considering the nature of high dimension using the common bag-of-words (BoW) representation, where the number of word features is ultra large and exceeds the number of documents, direct estimation of logistic regression may fail because the matrix inversion can be singular, besides other numerical problems that result in a lack of convergence, overfitting, and poor predictive accuracy, see for example \cite{Greenland2000} and \cite{Hadjicostas2003}. In view of these constraints, a sparsity threshold is usually applied before the use of logistic regression to retain only a portion of original features and truncate away sparse ones. However, the disadvantage of the truncated logistic regression is that some potentially important features may get filtered out, thus having no chance of entering model estimation stage for fair comparison. Meanwhile, classifiers such as SVM employ non-linear transformation of the original feature space, which in general gives an improved predictive performance. Nevertheless, these methods often give a black-box solution that makes it difficult to interpret feature importance, see \cite{Francis2006}.
	
	From a statistical viewpoint, the challenge of high dimensionality can be handled using regularization, which is often used to perform variables selection in regression under sparsity assumption. Instead of thinking that all the word features are relevant, we assume that only a small number of features dominate the sentiment classification and are therefore statistically significant. Various penalty functions have been proposed to facilitate sparsity by shrinking the coefficients of non-significant features to zero. The literature of sparsity is still growing, see for example LASSO \cite{Tibshirani1996}, SCAD \cite{Fan2001} and MCP \cite{Zhang2010}. In the literature of sentiment analysis, regularization has been applied in linguistic structure \cite{Yogatama2014}, contextual knowledge \cite{Wu2015} and graph representation \cite{Dai2015}. In addition, \cite{Genkin2007} assess the effect of regularization in sentiment classification using a Laplace prior. Nevertheless, none of the existing literature explores the advantage of regularization in deriving managerial insight. A related regularization based approach on topic modelling is discussed in \cite{topic2016}, where the differential use of word frequency is highlighted in characterizing topical content.
	
	In our study, we propose a Regularized Text Logistic (RTL) regression model to simultaneously perform sentiment classification and identify important word features from vast numbers of textual reviews. We develop a penalized local maximum likelihood estimator using the SCAD penalty, and establish the consistency and oracle property of the estimator in the RTL framework. We also extend the local estimator to global estimator and show the global optimum can be obtained under mild conditions with the number of reviews and features both diverging. A coordinate descent algorithm is used to solve for the global optimum of the penalized likelihood function with nonconvex penalty. We apply the RTL model to two online review datasets, Restaurant and Hotel, from TripAdvisor. Our results demonstrate satisfactory classification performance compared with alternative classifiers, with a highest true positive rate of 94.9\%. Moreover, RTL identifies a small set of word features, corresponding to 3\% for Restaurant and 20\% for Hotel, which boosts working efficiency by allowing managers to drill down into a much smaller set of important customer reviews. The selescted word features are further categorized into five groups for more targeted operational analysis, namely: non-informative adjectives or nouns, food or room quality, service, brand recognition, and view. 
	
	The rest of the paper is structured as follows. Section 2 describes the data used in our analysis. Section 3 introduces the proposed RTL model. Section 4 focuses on real data analysis by comparing with other alternative classifiers. Finally, Section 5 concludes this paper.

	\section{Data}\label{sec:DataAndMethodology}
	
	We consider two datasets of tourist reviews on restaurants and hotels that were posted on TripAdvisor. The Restaurant dataset contains 1,899 English reviews on nine major restaurants in a world-famous hotel located in Singapore from 1 December 2015 till 9 September 2016. The Hotel dataset includes 2,519 English reviews on hotels in different cities in the United States from 28 March 2001 to 27 January 2009. The Hotel dataset is part of the benchmark data in hospitality industry research that was used in \cite{Wang2011}. Each review contains attributes such as date, title, full text, date and customer satisfaction rating score (1--5). In our study, a review is classified as Positive if the rating score is 4--5, and Negative if the rating score is 1--2, and the ratings with a score 3 were discarded.
	
	Two sample reviews from the Restaurant dataset follow. The first is a positive review with score 5, where the customer shared his satisfaction with view, food, wine, service. The second is a negative review with score 1, where the customer expressed his disappointment in many things.

	\textit{ ``We sat outside overlooking the promenade, superb food, wines and service. Try the traditional fish and chips, and lamb chops!
		Lucky to get a table as we hadn’t made a booking. Lovely complimentary breads and with my hungry adult males they
		brought out more while waiting on our mains! Nice touch!" (posted
		on 8 December 2015)} \\
	
	\textit{ ``Came here on Sunday night with friends with high expectations
		because of the famous chef behind the brand, I could not be more
		disappointed in terms of food or service, I suppose coming from
		Europe I am use to good French bistro with good standard but
		here it’s fine dining with Michelin star price ouch!'' (posted on 21
		March 2016)}

	Compared to the reviews on restaurants posted in a relatively recent time, the hotel reviews posted in much earlier years are on average shorter in length. The variation is possibly accompanied with the change of customers' acceptance and behaviors in social medias. Two sample reviews from the Hotel dataset follow, with positive and negative scores, respectively:

	\textit{  ``Quality hotel at great price. Very clean. Free breakfast with good selections. Staff friendly and most helpful. A great stay!"
		(posted on 25 November 2008)} \\
	
	\textit{  ``Don’t do it!! This place is run down, dirty and loud. The
		pictures they provide on the web do not tell the story so don’t be
		fooled.'' (posted on 6 March 2008)}

	We split the sample into training corpus with 80\% of total reviews, including validation samples depending on the number of folds used in cross validation. The other 20\% are used as testing corpus for out-of-sample assessment. The training sample is used to estimate the unknown parameters in the sentiment classifications. The validation set helps to learn the hyper-parameters in the classifiers for extracting key words. The testing sample is reserved for out-of-sample prediction experiments.

	The objectives in our study are: 1) to build a classifier which can accurately classify the polarity of a review as positive or negative given its textual content, and 2) to identify meaningful word features that can be used for polarity interpretation and managerial improvement. To facilitate quantitative analysis, it is necessary to transform the unstructured free form textual reviews into a corpus in a structured form. The corpus is represented as a document-term matrix, where each entry represents a word feature in a document. We follow the standard pre-processing process to retain and arrange meaningful words and to filter out uninformative ones. Compared with direct transformation, this alleviates challenges in both memory and computation by avoiding an unnecessarily large matrix. The pre-processing steps include: (1) transform into lowercase, (2) remove punctuation, (3) remove stop words (e.g. the, a), (4) strip out white space, and (5) stem words to remove suffixes (i.e., services, server, and served are all replaced by serv).

	Table \ref{tbl:review_summary} gives the summary statistics of the two corpora. Among the 2,519 reviews on hotels, there are in total 1,559 positive reviews and 600 negatives, containing 11,324 features (i.e. unique words). Similarly, there are 1,527 positive and 130 negative reviews on the restaurants with 5,543 features. Note that this is a typical ultra-high dimensional problem in statistical learning with large ${p}/{n}$ ratio, where $p$ is the number of features and $n$ is the number of training examples.

	\begin{table}[htbp]
		\begin{center}
			\caption{Summary statistics of the datasets}\label{tbl:review_summary}
			\begin{tabular}
				{p{2cm}p{2cm}p{2cm}p{2cm}p{2cm}}
				
				\hline\hline
				{Dataset} & {Region} & {Positive} & {Negative} & {Word features} \\
				
				\hline
				Hotel & US & 1559 & 600 & 11324
				\\
				
				Restaurant & Singapore & 1527 & 130 & 5543   \\
				\hline\hline    
				
			\end{tabular}
		\end{center}
	\end{table}

	The most popular transformation is bag-of-words (BoW) representation, which assumes that the distribution of words within each corpus is sufficient and that linguistic features such as order and grammar can be safely ignored for sentiment analysis. In BoW representation, reviews are transformed into a document-term matrix, which contains a column for each word appearing in the corpus and a row for each review. The document term matrix often contains a huge number of columns, which corresponds to the number of the unique word features. If each matrix entry is the count that indicates the number of times a particular word feature appears in each review, then it is referred to as frequency statistic. However, this frequency statistic ignores the fact that some words in general appear more frequently than others. We adopt the tf-idf statistic (short for term frequency - inverse document frequency) to indicate the importance of a word in the corpus. 
	
	Let $\Omega=\{\omega_1, w_2,..., \omega_{p_n}\}$ denote the word features in the corpus, $d=\{d_1, d_2, ...d_n\}$ the set of reviews, and $x_{ij}$ the tf-idf weight of feature $w_j$ in the review $d_i$. The tf-idf for each term $x_{ij}$ is calculated as the product of two statistics, term frequency (tf) and inverse document frequency (idf), where 
	\begin{eqnarray*}
		\mbox{tf}(x_j) &=& \frac{\#x_j \mbox{ appears in review } d_i}{\#\mbox{all terms in review } d_i},\\
		\mbox{idf}(x) &=& \ln( \sum^n_{i=1}d_i)-\ln (\sum^n_{i=1}d_iI(x_j \in d_i)).\\ 
	\end{eqnarray*}
	In tf-idf, the word importance increases proportionally to the number of times it appears in the document but is offset by the its frequency in the corpus. The td-idf statistic has become one of the most popular feature statistics in text analysis, and 83\% of text-based recommender systems in the domain of digital libraries use tf-idf, see \cite{survey_2015}.

	\section{Regularized Text Logistic Regression}\label{sec:method}
	
	In this section, we present Regularized Text Logistic (RTL) regression to perform sentiment classification on text reviews. RTL assumes a sparse structure in the logistic regression where only a small number of word features are dominating and therefore can be used for ultra-high dimensional analysis with relatively small sample size. We conduct the estimation with penalized maximum likelihood and derive the asymptotic properties of the estimator. 
	
	Like the classic logistic regression, RTL models the conditional probability of the response taking a particular value. Meanwhile, RTL is able to efficiently perform a supervised classification given the high-dimensional word features in the BoW representation. In contrast, a direct application of the classic logistic regression suffers from numerical problems such as lack of convergence, overfitting, and poor predictive accuracy, due to the curse of dimensionality, especially when the number of word features exceeds the number of documents. By enforcing sparsity on the features, RTL not only builds a statistical classifier with good predictive power but also helps to identify key word features that can be used for polarity interpretation and managerial improvement.
	
	Recall that $x=\{x_1, x_2,..., x_{p_n}\}$ is the word features represented by the tf-idf statistic, and $d=\{d_1, d_2, ...d_n\}$ is the set of reviews. The RTL model is defined by the following process:
	\begin{equation}
	y_i = f(x_i^T\beta_n)+\epsilon_i
	\end{equation}
	where $y_i$ takes the value of 1 if the $i-$th review $d_i$ is positive, and 0 if negative. The word features $x_i$ is a vector of tf-idf statistic, and $f$ is the sigmoid link function used in the logistic regression. For the binary response, the conditional probability is 
	$$P(y_i=1|x_{i1},...,x_{ip_n})=  \pi_i=\frac{e^{x_i\beta_n}}{1+e^{x_i\beta_n}}$$
	where $\{\pi_i,i=1...n\}$ is predicted conditional probability of the binary response being 1 given $\{x_i,i=1...n\}$, and $\beta_n$ is the unknown coefficient vector. Under sparsity, only a subset of features is supposed to have significant impact in determining review polarity, while the other coefficients are zero. Denote ${\beta_n=(\beta_{1n}^T,\beta_{2n}^T)^T}$ as the set of coefficients, where $\beta_{1n}$ contains the non-zero coefficients, and $\beta_{2n}$ is the set of zeros. The optimal classifier can be obtained by minimizing the negative log-likelihood function 
	$$l_n(\beta_n)=-\frac{1}{n}\sum\limits_{i=1}^{n}\{y_ilog(\pi_i)+(1-y_i)log(1-\pi_i)\}$$
	with penalty on word features:
	$$Q_n(\beta_n) = l_n(\beta_n)+ h_\rho(\beta_n)$$
	where $h_\rho(\beta_n)$ represents a penalty function on the covariates $\beta_n$ and $\rho$ denotes the tuning parameters.

\begin{figure}%
		\centering
		{{\includegraphics[width=12cm]{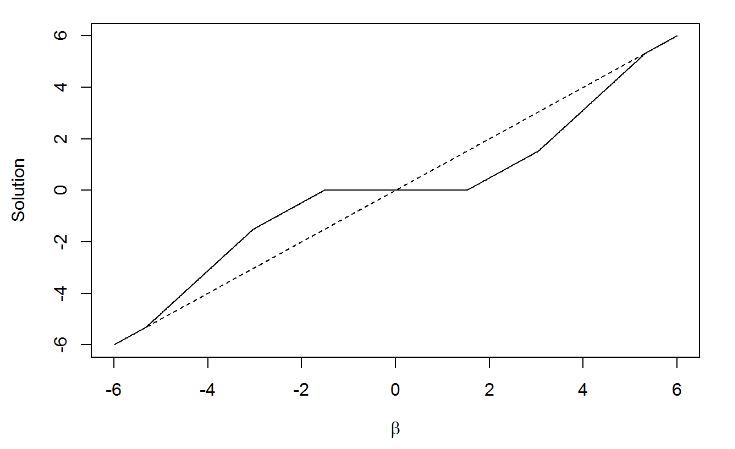} }}%
		\caption{SCAD thresholding rule}%
		\label{fig:scad}%
	\end{figure}

	Various penalty functions have been proposed to facilitate sparsity by shrinking the coefficients of non-significant features to zeros.  \cite{Genkin2007} adopts the $\ell_1$ type penalty in text categorization, which is shown to be equivalent to using a Laplace prior in a Bayesian logistic regression framework. However, the $\ell_1$ penalty does not possess the three theoretical properties of sparse estimators defined in \cite{Fan2001}, namely: unbiasedness, sparsity and continuity. Moreover, there is a diverging problem as new word features occur when the sample size gets large (i.e. web data where $p_n$ is diverging together with $n$), see \cite{Donoho2000}. \cite{Fan2004} extended the Smoothly Clipped Absolute Deviation (SCAD) regularization framework to the case of diverging number of parameters. \cite{Wang2016} proposed the estimation using quadratic approximation and proved the existence of global optimum with mild conditions. This motivates the adoption of the SCAD penalty in the sentiment classification. We have
	\begin{equation}
	h_\rho(\beta_n)=\left\{
	\begin{array}{ll}
	\lambda_n|\beta_j|   \qquad \qquad \qquad if \quad |\beta_j|\leq\lambda_n;\\
	-(\frac{|\beta_j|^2-2\gamma\lambda_n|\beta_j|+\lambda_n^2}{2(\gamma-1)}) \qquad if \quad \lambda_n<|\beta_j|\leq\gamma\lambda_n;\\
	\frac{(\gamma+1)\lambda_n^2}{2} \qquad \qquad if \quad |\beta_j| > \gamma\lambda_n
	\end{array}
	\right.
	\end{equation}
	where $\rho = (\lambda_n,\gamma)$  are hyper tuning parameters. The penalty function forms a quadratic spline function with knots at $\lambda_n$ and $\gamma\lambda_n$. It is continuously differentiable on $(-\infty,0)\cup(0,\infty)$ but singular at 0 with its derivatives being 0 outside the range $[-\gamma\lambda_n,\gamma\lambda_n]$. Under this regularization mechanism, the small coefficients are set to 0, a few other coefficients are shrunk towards 0, while the large coefficients are retained as they are, thus producing sparse solution and approximately unbiased estimates for the large coefficients. Taking the first derivative for some $\gamma>2$ and $\beta_n>0$, we have
	$$h\prime_\rho(\beta_n)=\lambda_n\{I(\beta_n \leq \lambda_n)+\frac{(\gamma\lambda_n-\beta_n)_+}{(\gamma-1)\lambda_n}I(\beta_n>\lambda_n)\},$$
	which shows that the penalty initially applies the same rate of penalization as LASSO \cite{Tibshirani1996} and gradually reduces to 0 as $\beta_n$ gets larger. As shown in Figure \ref{fig:scad}, the solution with the SCAD penalty function results in soft thresholding rule:
	\[
	\hat\beta_j^{SCAD}=\left\{
	\begin{array}{ll}
	(|\hat\beta_j|-\lambda_n)_+\text{sign}(\hat\beta_j)   \qquad \qquad \quad \quad  \text{if}   \quad |\hat\beta_j|<2\lambda_n;\\
	\{(\gamma-1)\hat\beta_j-\text{sign}(\hat\beta_j)\gamma\lambda_n\}/(\gamma-2)
	\quad \text{if}  \quad 2\lambda_n<|\hat\beta_j|\leq\gamma\lambda_n;\\
	
	\hat\beta_j \qquad \qquad \qquad \qquad \qquad \qquad \qquad \text{if} \quad |\hat\beta_j| > \gamma\lambda_n
	\end{array}
	\right.
	\]
	where $\hat\beta_j$ is the estimator without regularization. The hyper parameters $\lambda_n$ and $\gamma$ can be obtained via two dimensional grid search using criteria such as cross validation, generalized cross validation, AIC, and BIC, see \cite{survey2018}.
	
	\subsection{Numerical Algorithm}\label{sec:algo}
	We employ the efficient coordinate descent algorithm \cite{Breheny2011}. In particular, the coordinate descent algorithm partially optimizes $Q_n(\beta_n)$ with respect to a single parameter $\{\beta_j,j=1,...,p_n\}$ at a time with the remaining elements being fixed at their most recent updated values. The procedure iteratively repeats through all the parameters until convergence is reached. In the context of RTL regression, quadratic approximation to the loss function is used based on a Taylor series expansion, which results in iteratively reweighted least squares algorithm commonly used in the generalized linear models. Specifically, we have
	\begin{equation}
	\label{eq:1}
	{Q_n(\beta_n) \approx \frac{1}{2n}(\tilde{y}-X\beta_n)^TW(\tilde{y}-X\beta_n)+\sum\limits^{p_n}_{j=1}h_{\rho}(|\beta_j|)}
	\end{equation}
	where $\tilde{y}$ is the working response
	\begin{equation}
	{\tilde{y}=X\beta^{(m)}+W^{-1}(y-\pi)}
	\end{equation}
	and ${W}$ is a diagonal matrix of weights with elements $w_i=\pi_i(1-\pi_i)$, and ${\pi_i}$ is evaluated at $\beta^{(m)}$.
	
	The algorithm goes as follows. At iteration $m$, approximate the loss function via (\ref{eq:1}) given $\beta^{(m)}$, and execute the following calculations to obtain $\beta^{(m+1)}$ for each feature $\{x_j, j =1,..., p_n\}$: 
	\begin{enumerate}
		\item calculate the partial derivative of the unpenalized regression on $x_{j}$, denoted as 
		$$
		z_j=n^{-1}x'_{j}W(\tilde{y}-X_{-j}\beta_{-j})=n^{-1}x'_{j}Wr+v_j\beta_j^{(m)}
		$$
		where $-j$ refers to the portion that remains after removing the $j^{th}$ feature, $\beta_{-j}^{(m)}$is the most recently updated value of $\beta$, $r=W^{-1}(\tilde{y}-\pi)$ is current residual, and $v_j=n^{-1}x'_{j}Wx_j$. \\
		\item update $\beta_{j}^{(m+1)} \xleftarrow[]{} h_{\rho}(z_j)$
		where 
		\[
		\hat\beta_j^{SCAD}=h_{\rho}(z_j)=
		\left\{
		\begin{array}{ll}
		\frac{S(z_j,\lambda)}{v_j}  \qquad \qquad \quad \quad  \quad\text{if}   \quad |z_j|\leq \lambda_n(v_j+1);\\
		\frac{S(z_j,\gamma\lambda_n/(\gamma-1))}{v_j-1/(\gamma-1)}
		\quad \quad  \qquad \text{if} \quad \lambda_n (v_j+1) < |z_j|\leq v_j\gamma\lambda_n;\\
		
		\frac{z_j}{v_j} \quad \quad \quad  \qquad \qquad \qquad \text{if} \quad |z_j| > v_j\gamma\lambda_n
		\end{array}
		\right.
		\]
		\item update $r \xleftarrow[]{} r-W^{-1}(\beta_j^{(m+1)}-\beta_j^{(m)})x_j $.
	\end{enumerate}
	To achieve a continuous and stable solution, we adopt the adaptive rescaling of the penalty parameter $\gamma$ \cite{Breheny2011} to match the continually changing scale of the covariates during training.
	
	The optimal $\lambda_n$ and $\gamma$ are selected in a data-adaptive approach by minimizing the prediction errors in a $K$-th folder cross validation:
	\[
	CV(\hat y^{(k)})=\frac{1}{N}\sum^N_{i=1}Q_n(y_i,\hat y^{(-k)})
	\]
	where $\hat y^{(-k)}$ is the fitted function with the $k^{th}$ subset removed from the original corpus for $k=1,2,...,K$. There are three hyper-parameters: the number of folds $K$ in cross validation, the tuning parameter $\gamma$, and penalty coefficient $\lambda_n$. We allow $K$, $\gamma$ and $\lambda_n$ to vary jointly along a 3D grid with $K$ ranging from 5 to 20, $\gamma$ from 2.1 to 4 with a step size of 0.1, and $\lambda_n$ from a range of values using a warm start  in the search of minimizing the cross validation errors.

	\subsection{Asymptotic Properties }\label{sec:theory}
	We study the sampling properties of the proposed penalized likelihood estimator with diverging dimensionality and discuss the required regularity conditions to obtain the global minimizer. Let ${\beta_0=(\beta_{10}^T,\beta_{20}^T)^T}$ be the true set of coefficients to be estimated, where $\beta_{10}$ is a $k_n$x1 vector of significant coefficients and $\beta_{20}$ is an $m_n$x1 vector of non-significant coefficients satisfying $k_n+m_n=p_n$. Recall that the total number of parameters $p_n$ is allowed to grow slowly as sample size goes to infinity. Without loss of generality, we assume that $\beta_{20}=0$. Let $\tilde{\beta}_n$ denote the global minimizer and $\hat{\beta}_n$ the proposed RTL estimator. The regularity conditions required to establish the asymptotic properties are provided in the appendix.

	\begin{theorem}
		Suppose that Conditions (C.1), (C.3), (C.4) and (C.7) hold, then there exists a local minimizer $\hat{\beta}_n=(\hat{\beta}_{1n}^T,\hat{\beta}_{2n}^T)^T$ of $Q_n(\beta_n)$ such that 
		\begin{center}
			$\Vert\hat{\beta}_n-\beta_0\Vert=O_p(\sqrt{p_n/n})$
		\end{center}
	\end{theorem}
	
	\noindent This shows that under some conditions, there exists a root-($p_n/n$)-consistent estimator. The following lemma provides sparsity.
	
	\begin{lemma}
		Suppose (C.1)-(C.7) hold. Then the SCAD estimator $\hat{\beta}_n$ satisfies
		\begin{center}
			$Pr(\hat{\beta}_{2n}=0) \rightarrow 1$
		\end{center}
	\end{lemma}
	
	\noindent Moreover, we can show the oracle property of the estimator. Denote
	
	\begin{align*}
		\Sigma_{\lambda_n}&=\{h\prime\prime_{\lambda_n}(|\beta_{01}|),...,h\prime\prime_{\lambda_n}(|\beta_{0k_n}|)\} \\
		b_{_n}&=\{h\prime_{\lambda_n}(|\beta_{01}|)\mbox{sgn}(\beta_{01}),...,h\prime_{\lambda_n}(|\beta_{0k_n}|)\mbox{sgn}(\beta_{0k_n})\}^T 
	\end{align*}
	where diag$\{\cdot\}$ is a diagonal matrix and sng$(\cdot)$ is a sign function.

	\begin{theorem}
		Suppose that $p^2_n/n \rightarrow 0$, and the regularity conditions (C.1)-(C.7) are satisfied, then the local minimizer $\hat{\beta}_n=(\hat{\beta}_{1n}^T,\hat{\beta}_{2n}^T)^T$ of $Q_n(\beta_n)$ in Theorem 1 satisfies \\
		(1) Sparsity: $Pr(\hat{\beta}_{2n}=0) \rightarrow 1$ as $n \rightarrow \infty$. \\
		(2) Asymptotic normality:
		$$\sqrt{n}\alpha^T_nI_{n}^{-1/2}(\beta_{10})(I_{n}(\beta_{10})+\Sigma_{\lambda_n})[(\hat{\beta}_{1n}-\beta_{10})+(I_{n}(\beta_{10})+\Sigma_{\lambda_n})^{-1}b_n] \xrightarrow{D} N(0,1)$$
		where $\alpha_n$ is an arbitrary $k_n$\textup{x}1 vector such that $\Vert\alpha_n\Vert=1$.
	\end{theorem}
	This theorem ensures model sparsity and asymptotic normality when the number of parameters diverges. When $n$ is large enough, it holds that $\Sigma_{\lambda_n}=0$ and $b_n=0$ for SCAD penalty, and the asymptotic normality becomes
	$$\sqrt{n}\alpha^T_n I_{n}^{1/2}(\beta_{10})(\hat{\beta}_{1n}-\beta_{10}) \xrightarrow{D} N(0,1)$$
	which is as efficient as the maximum likelihood estimator of $\beta_{10}$ if $\beta_{20}$ were known in advance. We also show the global property of the estimator
	\begin{theorem}
		Under conditions (C.1), (C.5) and (C.7), with probability tending to 1, the local minimizer $\hat{\beta}_n$ is the global minimizer of (1); that is,
		$$Pr(Q_n(\hat{\beta}_n)=inf_{\beta_n}Q_n(\beta_n)) \rightarrow 1$$
	\end{theorem}

	\section{Real Data Analysis}\label{sec:Result}
	
	We implement the proposed RTL regression on the Restaurant and Hotel datasets. Our primary objective is to identify a set of key features amenable for operational improvement.  Along with the real data analysis, we  also  investigate the sentiment classification performance of the proposed RTL approach and compare with several popular alternative sentiment classification approaches. Our code and data are available on Github\footnote{https://github.com/jackliu333/regularized-text-logistic-regression.}.

	\subsection{Evaluation Measure}\label{sec:PerformanceMeasure}
	In general, an accurate classifier is characterized by correct sentiment classifications. There is usually a preference between precision and recall, depending on the business case. For example, hoteliers in the service industry may not want to miss out any actual negative reviews and thus may pay more attention to false negatives than false positives. To measure the performance of sentiment classifier from different angles, we consider six commonly used metrics in the literature.  The first four metrics---namely sensitivity, specificity, precision, and negative predictive value---target  a specific aspect of interest, while the last two---namely accuracy and F1 score---are overall scores on correct classification and positive polarity, respectively. See Table \ref{tbl:metrics} for the definition and interpretation.
	
	\begin{table}[]
		\centering
		\resizebox{1\textwidth}{!}{\begin{minipage}{\textwidth}
				\caption{Six metrics to measure model performance. Letter $a$ counts the documents correctly classified as positive reviews; $b$ counts the documents incorrectly classified as positive reviews; $c$ counts the documents incorrectly classified as negative reviews; and $d$ counts the documents correctly classified as negative reviews.}\label{tbl:metrics}
				\begin{tabular}{p{3cm}|p{3cm}|p{7cm}}
					\hline\hline
					Metric & Definition & Interpretation \\ \hline
					True positive rate (TPR) & \centering$\frac{a}{a+c}$ & Also called recall or sensitivity, measures the proportion of positive reviews that are correctly identified, or the extent to which true positives are not missed. \\ \hline
					True negative rate (TNR) & \centering$\frac{d}{b+d}$ & Also called specificity, measures the proportion of negatives that are correctly identified. \\ \hline
					Positive predictive value (PPV) & \centering$\frac{a}{a+b}$ & Also called precision, measures the proportion of predicted positive reviews that are true positive. \\ \hline
					Negative predictive value (NPV) & \centering$\frac{d}{c+d}$ & Measures the proportion of predicted negative reviews that are true negative. \\ \hline
					Accuracy & \centering$\frac{a+d}{a+b+c+d}$ & Measures the proportion of reviews that are correctly classified.
					\\ \hline
					F1 Score & 
					\centering$\frac{2 * PPV * TPR}{PPV + TPR}$ & Combines precision and recall as an overall metric
					\\ \hline
				\end{tabular}
		\end{minipage}}
	\end{table}

	\subsection{Alternative Classifiers}\label{sec:PerformanceMeasure}
	
	We consider three machine learning classifiers as alternatives, namely: Naive Bayes (NB), K-Nearest Neighbor (KNN) and Support Vector Machine (SVM). The classifiers usually provide good accuracy, although they are incapable of providing key words in the textual reviews. We compare the performance of RTL on the original document-term matrix together with the alternatives, with the purpose of testing the performance of RTL against other benchmark methods in high dimensional setting. 
	
	\noindent\textbf{NB}: The NB classifier is developed from Bayesian probability and assumes that the conditional probabilities of word features are independent of each other, see \cite{NB_2014}. The joint distribution of review sentiment $y_i$ depending on the word features $\{x_{1}...x_{p_n}\}$ is 
	\begin{equation}
	Pr(y_i,x_{1}...x_{p_n})=Pr(y_i) \cdot \prod^{p_n}_{j=1}Pr(x_{j}|y_i)
	\end{equation}
	where $Pr(y_i)$ is the sentiment prior and $Pr(x_j|y_i)$ denote the conditional distributions. NB can be estimated using either maximum likelihood or maximum a posteriori approach. The polarity of a new review $y_i^*$ is predicted by choosing the category with the maximum posterior probability given the corresponding word features. 
	\begin{equation}
	y_i^*=argmax_{y_i}Pr(y_i|x_{1}...x_{p_n})
	\end{equation}
	
	\noindent\textbf{SVM}: The SVM classifier identifies the linear or non-linear separator that can discriminate different classes in the search space. Training the classifier involves minimizing the error function
	\begin{equation}
	\begin{split}
	\frac{1}{2}\beta^T\beta+C\sum^N_{i=1}\xi_i \\
	s.t. \quad    y_i(\beta^T\phi(x_i)+b) \geq 1-\xi_i \\
	\xi_i \geq 0, i=1,...N
	\end{split}
	\end{equation}
	where $C$ and $b$ are constants, $\xi_i$ denotes parameters for handling non-separable data, and $\phi$ is the kernel function used to transform the data to intended feature space. 
	
	\noindent\textbf{KNN}: The k-nearest neighbors algorithm is a non-parametric approach by a majority vote of its neighbors for sentiment classification. In other words, a review is assigned to the most common polarity among its k nearest neighbors using Euclidean distance after normalizing the feature weights. 
	
	The machine learning classifiers rely on the choice of hyper-parameters. We follow the same data-adaptive approach and select the respective hyper-parameters via cross validation. For example, the cost is allowed to vary from a grid of values between $10^{-1}$ and $10^2$, $\gamma$ among 0.5, 1 and 2 for SVM with a linear kernel. 
	
	Moreover, we also consider the implementation of the truncated Logistic Regression (LR), with pre-determined ``key" word features. In particular, we sort the word features according to frequency and apply a number of thresholds to the document-term matrix. Given a certain sparsity threshold, the word features with lower frequency will be removed to reduce the  original feature space before applying logistic regression. A lower threshold will retain more features in the corpus. In the comparison, a range of threshold values from 0.8 to 0.99 with a step size of 0.01 is used. In addition to accuracy, we are interested in the difference of the selected features by an automatic approach using RTL compared to those pre-determined features in the truncated LRs.

	\subsection{Word Features}
	
\begin{figure}
		\begin{center}
			\includegraphics[width=14cm, height=10cm]{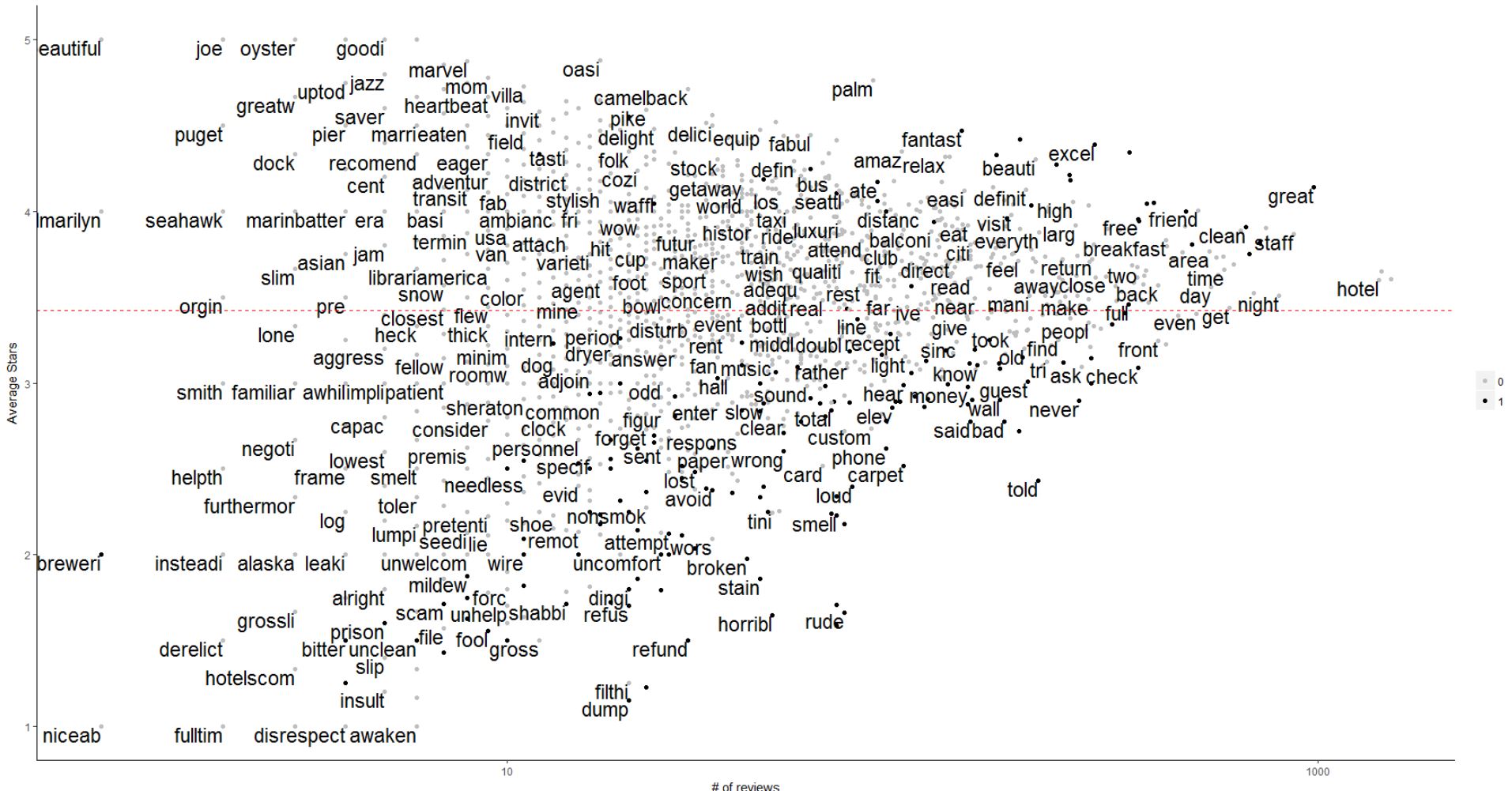}\hspace{0.1cm}
			\caption{Selected words of Hotel dataset}\label{fig:word_distribution_scad_hotel}
		\end{center}
	\end{figure}

\begin{figure}
		\begin{center}
			\includegraphics[width=14cm, height=10cm]{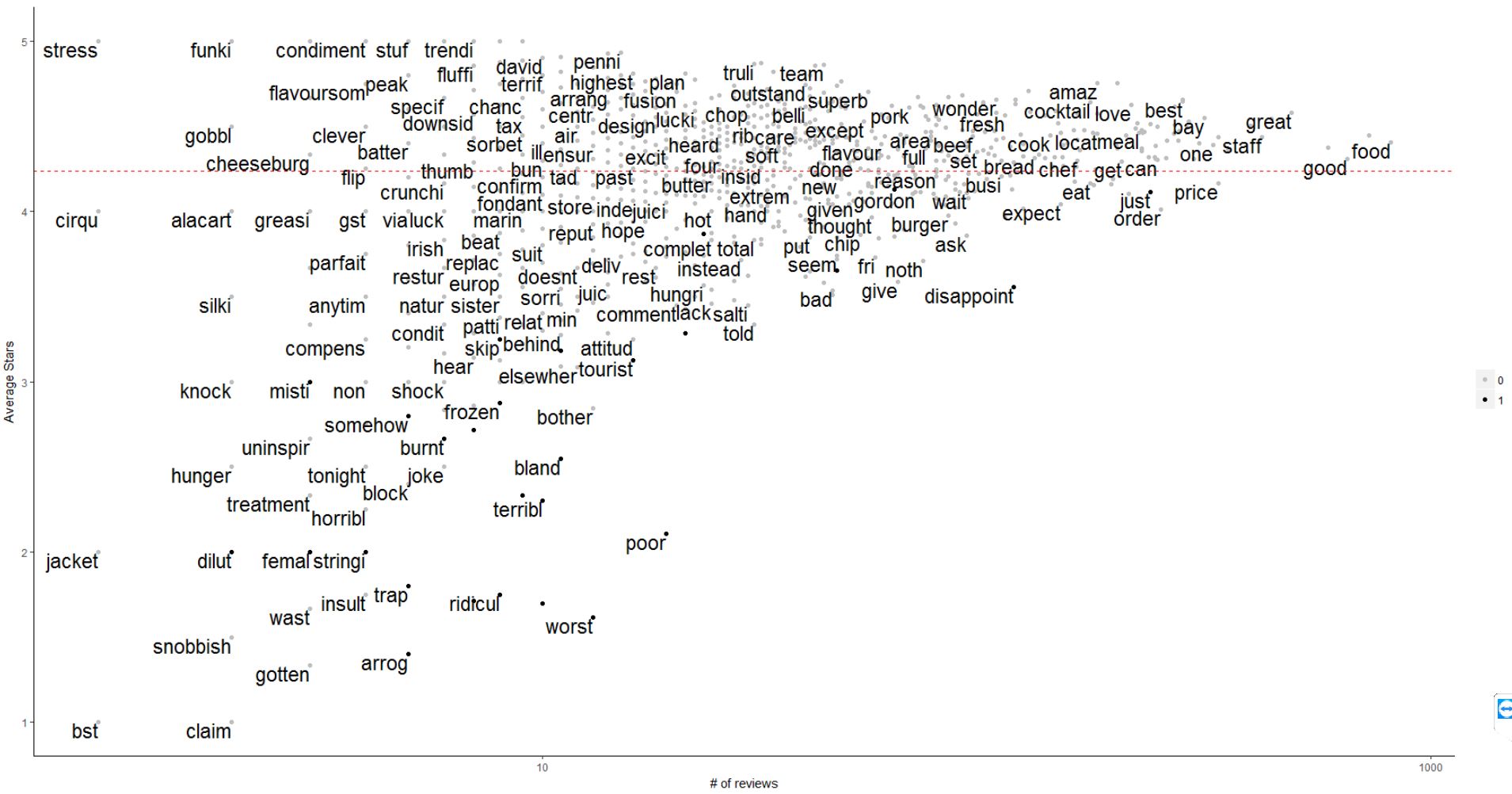}\hspace{0.1cm}
			\caption{Selected words of Restaurant dataset}\label{fig:word_distribution_scad_restaurant}
		\end{center}
	\end{figure}

	There are 357 words out of 11,324 (Hotel) and 115 out of 5,543 (Restaurant) features selected by RTL from the two datasets, corresponding to 3\% and 20\% of the original word features, respectively.  Figure \ref{fig:word_distribution_scad_hotel} \& \ref{fig:word_distribution_scad_restaurant} display the selected features, labelled with the words, where the insignificant features are only denoted as dots. The average rating is represented as a red line. In both cases, the average score is positive. Some low frequency words are scattered on the left side of the distribution, while the majority are concentrated in the middle. It is interesting to note that common words such as ``great" and ``bad" express little
	information on the drivers of sentiment orientation. Meanwhile, rare words with low frequency like ``stringi"
	and ``dilut" are also selected by the RTL model, which are more self-explanatory and indicative of the key points in a review.

	The words in the original feature space are generally more informative than transformed ones. For example, ``diluted'' and ``stingy'' express the potential pain points on food and beverages, which could assist restaurant
	managers in drilling down into specific reviews for detailed analysis. This can also be used to build up hotel and restaurant management lexicon for text analytics. In addition, working efficiency can also be improved in terms of the number of reviews needed for perusal. By focusing on the reviews containing at least one of the selected features, managers only need to focus on a much smaller set of relevant reviews to understand the action points, instead of browsing through a large but not necessarily informative set of reviews.
	
	Most selected features fall into the lower half of the word distribution plot, including those that occur only a few times. It should be noted that all of the selected features in the Restaurant dataset have a negative coefficient, while in the Hotel dataset there are some positive features selected, although negative features still account for the majority. This is possibly due to the small sample size of the Restaurant dataset. In the case of high dimensional text mining, when the dataset is relatively small, only those features with high weightage, though few occurrence, stand out, while the majority of the features will be penalized by the strict regularization term during the model estimation stage. 
	
	Among the selected features, the negative features have bigger coefficients and in general less frequency than positive features. All features appear to have relatively high review length on average, suggesting that the feature coefficient is positively correlated with review length and negatively correlated with its frequency. For example, among the six reviews that contain "macdonald", which has the most negative coefficient, there are 168,533 characters and 458 word features in negative reviews, and 17,447 and 675 in positive reviews respectively. Similarly, the only review that contains "airhotelrent" has 332 word features, indicating the importance of review length along with its tf-idf weights in determining feature coefficient. The argument for other features is almost the same, except for a few less informative features such as "poor" or "disappoint". Features with positive coefficients are in general associated with much higher frequency than in negative reviews. 
	
	There is a discrepancy between the selected features from RTL and those by the truncated LR models. As shown in table \ref{T:feature_comparison}, besides 128 common features which are selected by both RTL and the truncated LR models at 88\% for Hotel Dataset\footnote{We selected this threshold for comparison as it gives highest F1 score among all the alternatives}, an additional 229 selected features are unique in the RTL model, which is contributed to the effect of regularization and could be perceived as the differentiating power compared with other alternative LR classifiers.
	
	To gain a better understanding of those selected features from RTL model and seek for operational insights, we grouped them into five  categories, namely: non-informative adjectives or nouns, food or room quality, service, brand recognition, and view. Many features contain little value on indicating certain operational aspect of interest and thus fall into the first category. However, other features are able to provide insights on the potential  operational aspects mentioned in the review. For example, despite a few overlapping word features that are mostly non-informative nouns or adjectives, RTL clearly discovers more features on the other informative categories, including food or room quality, service, brand recognition and view. Specifically, 49 selected features from the Hotel dataset using RTL are grouped into food or room quality. Features such as "bedspillow" and "bedspread" under this group draw immediate attention to specific aspects on bed preparation mentioned in the review. In addition, "stringi" and "dilut" in the Restaurant dataset or "cigarett" and "circuit" in the Hotel dataset are both significant in determining the review polarity and also indicative of the key aspects covered in these reviews, thus effectively highlighting the challenges in the service. Meanwhile, features pre-determined by the truncated LR models are generally vague and are not indicative of specific points for operational improvement. See the appendix for a full list of selected features.
	
	%
	

	\begin{sidewaystable}[ph]
		
		\smallskip
		
		\centering
		\small
		\begin{tabular}{M{3cm}|M{2.5cm}M{2.5cm}M{2.5cm}|M{2.5cm}M{2.5cm}M{2.5cm}}
			\hline\hline
			\multicolumn{1}{c}{} &
			\multicolumn{3}{c}{Hotel} &
			\multicolumn{3}{c}{Restaurant} \\
			& RTL & Common & LR at 88\% & RTL & Common & LR at 91\%
			\\
			\hline 
			Unique features & 339 & 18 & 61 & 114 & 1 & 72 \\\hline
			Sample features & 
			chimney bay cigarett refurbish bedspillow bedspread drunken telephon   &
			dont check full book call got hollywood recommend   &
			room desk night ask door look stay small &
			bland dilut firewood pastri tasteful tenderloin undercook &
			just &
			amaz bay cocktail delici dessert dinner nice perfect \\\hline
			Non-informative adjectives or nouns & 
			262 & 17 & 57 & 95 & 1 & 64 \\\hline
			Food or room quality & 
			49 & 0 & 0 & 14 & 0 & 6 \\\hline
			Service & 
			13 & 0 & 2 & 1 & 0 & 0 \\\hline
			Brand recognition & 
			9 & 1 & 1 & 3 & 0 & 1 \\\hline
			View & 
			3 & 0 & 1 & 1 & 0 & 1 \\

			
			\hline\hline
		\end{tabular}
		\caption{Comparison of selected features from the RTL model and the best-performing truncated LR model. Despite a few overlapping word features that are mostly non-informative nouns or adjectives, RTL model clearly discovers more features on the other categories, including food or room quality, service, brand recognition and view. For example, 49 selected features from Hotel Dataset using RTL model are grouped into food or room quality. Features such as "bedspillow" and "bedspread" under this group draw immediate attention to specific aspects on bed preparation mentioned in the review.
		}
		\label{T:feature_comparison}
	\end{sidewaystable}

	The hyper-parameters of RTL are selected via cross validation. Specifically, we obtained an optimal value of $K=19$, $\gamma=4$ and $\lambda_n=0.014$ for the Hotel dataset, and  $K=13$, $\gamma=4$ and $\lambda_n=0.022$ for the Restaurant dataset. The common value of $\gamma=4$ shows that its optimal value should be selected using a data-adaptive approach along with other hyper-parameters in text classification setting, as opposed to the fixed $\gamma=3.7$ suggested by \cite{Fan2001}. In addition, an appropriate value of $\lambda_n$ is also selected using this approach for a proper trade-off between bias and variance.

	\subsection{Predictive Performance}\label{sec:Result}
	
	Table \ref{T:performance_all} reports the in-sample and out-of-sample classification performance of various classifiers for both the Hotel and Restaurant datasets. We found that the RTL model, in addition to identifying the key word features, in general demonstrates comparable performances over the alternative classifiers. It is very close to the best performing classifier at the training stage, with its TPR and F1 score being the second highest for both datasets. Besides, the effect of regularization is even more obvious when it comes to out-of-sample performance, where RTL demonstrates better performance than most of the alternative classifiers. For example, besides having the second highest scores in TPR, NPV and accuracy, its F1 score also stands the highest among all classifiers for the Restaurant dataset.

	\begin{sidewaystable}[ph]
		
		\smallskip
		
		\centering
		\small
		\begin{tabular}{c|ccccccc|cccccc}
			\hline
			\multicolumn{1}{c}{} &
			\multicolumn{7}{c}{Hotel} &
			\multicolumn{6}{c}{Restaurant} \\
			\multirow{2}{*}{} &
			\multirow{2}{*}{} &  \multirow{2}{*}{SVM} & \multirow{2}{*}{KNN} & \multirow{2}{*}{NB} & \multicolumn{2}{c}{LR$^*$} &
			\multirow{2}{*}{RTL} &
			\multirow{2}{*}{SVM} & \multirow{2}{*}{KNN} & \multirow{2}{*}{NB} & \multicolumn{2}{c}{LR$^*$} &
			\multirow{2}{*}{RTL}
			\\
			\hhline{~~~~~--~~~~--~}
			&
			& & & & LR$_{-}$(\%) & LR$^{+}$(\%) & 
			& & & & LR$_{-}$(\%) & LR$^{+}$(\%) &
			\\ \hline
			&
			TPR & .99 & .84 & \textbf{1.0} & .88(86) & .97(98) & .99 
			& \textbf{1.0} & \textbf{1.0} & \textbf{1.0} & .98(94) & \textbf{1.0}(96) & .99\\ 
			&
			TNR & \textbf{1.0} & .21 & 0 & .29(80) & .95(98) & .80  
			& \textbf{1.0} & 0 & 0 & .01(80) & \textbf{1.0(96)} & .26 \\ 
			&
			PPV & \textbf{1.0} & .74 & .72 & .78(90) & .98(98) & .93  
			& \textbf{1.0} & .92 & .92 & .92(80) & \textbf{1.0}(96) & .94  \\ 
			In &
			NPV & \textbf{.99} & .33 & 0 & .64(80) & .93(98) & .98 
			& \textbf{1.0} & NA & NA & .25(80) & \textbf{1.0}(96) & .96 
			\\ 
			&
			Accuracy & \textbf{.99} & .68 & .72 & .76(80) & .97(98) & .78 
			& \textbf{1.0} & .92 & .92 & .25(80) & \textbf{1.0}(96) & .94 
			\\
			& F1 & \textbf{.99} & .79 & .84 & .85(80) & .98(98) & .96
			& \textbf{1.0} & .96 & .96 & .96(80) & \textbf{1.0}(96) & .96
			
			\\

			\hline
			&
			TPR & .95 & .80 & \textbf{1.0} & .94(80) & .85(98) & .95 
			& \textbf{1.0} & \textbf{1.0} & \textbf{1.0} & .92(98) & \textbf{1.0}(80) & .99
			\\ 
			&
			TNR & \textbf{.73} & .17 & 0 & .28(80) & .66(88) & .51 
			& .12 & 0 & 0 & .01(80) & \textbf{.31}(90) & .12
			\\ 
			&
			PPV & \textbf{.90} & .72 & .72 & .77(80) & .87(88) & .84 
			& .93 & .92 & .92 & .92(80) & \textbf{.94}(90) & .93
			\\ 
			Out &
			NPV & \textbf{.85} & .24 & 0 & .60(98) & .70(92) & .79  
			& \textbf{1.0} & NA & NA & .18(99) & 1.0(86) & .75 \\
			& Accuracy & \textbf{.89} & .62 & .72 & .75(80) & .83(88) & .83 
			& .90 & .90 & .89 & .92(80) & \textbf{.94}(91) & .93
			\\
			&
			F1 & \textbf{.93} & .75 & .84 & .85(80) & .88(88) & .89 
			& \textbf{.96} & .95 & .95 & .89(99) & \textbf{.96}(91) & \textbf{.96}
			\\
			\hline
			&
			\# Features used & 11324 & 11324 & 11324 & 29 (80) & 1070(99) & 11324 
			& 5543 & 5543 & 5543 & 12(80) & 686(99) & 5543
			\\
			&
			\# Features selected & 11324 & 11324 & 11324 & 29(80) & 1070(99) & 357 
			& 5543 & 5543 & 5543 & 12(80) & 686(99) & 115
			\\

			
			\hline
		\end{tabular}
		\caption{(In) and (Out) classification performances of both Hotel and Restaurant datasets. NA is due to $c+d=0$, meaning no negative review has been predicted with the associated method. LR$^+$ and LR$_-$ display the best and worst performing truncated LR, respectively, with the corresponding sparsity level in the brackets. The best prediction results in all metrics are highlighted in bold. In general the RTL model shows good performance over alternative classifiers. For example, it stays very close to the best performing classifier at training stage, with its TPR and F1 score being the second highest for both datasets. Its out-of-sample performance is even better than most of the alternative classifiers. Besides being the second highest in TPR, NPV and accuracy, its F1 score stands the highest among all classifiers for Restaurant dataset. 
		}
		\label{T:performance_all}
	\end{sidewaystable}

	\section{Conclusion}\label{sec:Result}
	In this paper, we proposed a RTL regression model to analyze unstructured textual reviews. The RTL model shows potential in identifying a set of statistically significant and operationally insightful features that are further grouped into five categories, while it achieves satisfactory predictive performance compared with alternative classifiers, as represented by its highest F1 score of 0.96 in the Restaurant dataset and the second highest score of 0.89 in the Hotel dataset. Under the sparsity assumption, we derived the oracle properties of these selected features. From an efficiency perspective, the resulting set of reviews that contain these features are also much smaller, when compared with the original reviews, to be read through. Thus, the RTL model has the practical value to be applied in hospitality operations and it can act as an effective tool for understanding the true drivers of customer satisfaction. It should be noted that Deep Neural Networks (DNN) approaches have been widely used in recent years, which give promising results for sentiment classification tasks by treating each feature as word embedding and using complex architecture for training, such as Convolutional Neural Networks \cite{CNN_2014}.  Therefore, a potential future extension would be to combine DNN with sparsity constraints to further boost the predictive performance, while still obtaining a compact set of meaningful features.

	\begin{appendix}
		%
		%
		\section{Reviews Containing Informative Features}
		
		See Table \ref{tbl:informativeReviewRestaurant1} - \ref{tbl:informativeReviewHotel2} for sample reviews that contain informative word features.
		
		\begin{table}[]
			
			\centering
			\resizebox{1\textwidth}{!}{\begin{minipage}{\textwidth}
					\caption{Reviews containing the root stringi in Restaurant Dataset}\label{tbl:informativeReviewRestaurant1}
					\begin{tabular}{p{3cm}|p{11cm}}
						\hline\hline
						
						Rating & Content
						\\ \hline
						2 & The views were wonderful but really not worth the price...then a steak which was tough, \textbf{stringy}, and bland... \\ \hline
						2 & ...Steak had no sear and was very \textbf{stringy}...nothing more than overpriced pub food \\ \hline
						2 & The food is overpriced ...beef was \textbf{stringy}...They screwed up. \\ \hline
						2 & We went here with high expectations...there was red \textbf{stringy} veins through it...
						\\ 
						\hline \hline
						
					\end{tabular}
			\end{minipage}}
		\end{table}

		\begin{table}[]
			
			\centering
			\resizebox{1\textwidth}{!}{\begin{minipage}{\textwidth}
					\caption{Reviews containing the root dilut in Restaurant Dataset}\label{tbl:informativeReviewRestaurant2}
					\begin{tabular}{p{3cm}|p{11cm}}
						\hline\hline
						
						Rating & Content
						\\ \hline
						2 & ...did not have much taste \textbf{diluted}...\\ \hline
						2 & ...they look beautiful but are generally a little \textbf{diluted}...
						\\ 
						\hline \hline
						
					\end{tabular}
			\end{minipage}}
		\end{table}

		\begin{table}[]
			
			\centering
			\resizebox{1\textwidth}{!}{\begin{minipage}{\textwidth}
					\caption{Reviews containing the root cigarett in Hotel Dataset}\label{tbl:informativeReviewHotel1}
					\begin{tabular}{p{3cm}|p{11cm}}
						\hline\hline
						
						Rating & Content
						\\ \hline
						
						2 & ...Room was clean but was non-smoking and smelled like \textbf{cigarettes}. Wouldn't stay there again.\\ \hline
						2 & ...One nice touch was the fact that the bedroom window actually opened, which was a Godsend when the 'smoke free' hotel A/C system was pouring \textbf{cigarette} smoke into our room (ugh!) \\ \hline
						1 & ...\textbf{Cigarette} stains in the carpets. Dark and dingy... \\
						
						\hline\hline
					\end{tabular}
			\end{minipage}}
		\end{table}

		\begin{table}[]
			
			\centering
			\resizebox{1\textwidth}{!}{\begin{minipage}{\textwidth}
					\caption{Reviews containing the root circuit in Hotel Dataset}\label{tbl:informativeReviewHotel2}
					\begin{tabular}{p{3cm}|p{11cm}}
						\hline\hline
						
						Rating & Content
						\\ \hline
						2 & ...it was very noisy and I wouldn't have been able to sleep, had to call maintenance to turn off via a \textbf{circuit} breaker...
						\\ \hline
						2 & ...but I didn't have a noise problem, other than the deafeningly loud fan that I couldn't figure out how to shut off until I called the front desk the next day (hint: it was the giant switch in the \textbf{circuit} breaker)... \\
						
						\hline\hline
					\end{tabular}
			\end{minipage}}
		\end{table}

		\section{Theoretical Proof}
		
		\subsection{Regularity conditions}
		\begin{itemize}
			\item (C.1)$\Vert\tilde{\beta}_n-\beta_0\Vert^2=O_p(p_n/n)$. 
			
			\item (C.2) liminf$_{n \rightarrow \infty}$liminf$_{\beta_n \rightarrow 0^+}p\prime_{\lambda_n}(|\beta_{n}|)/\lambda_n > 0$.
			
			\item (C.3) $a_n=O(n^{-1/2})$ as $n \rightarrow \infty$.
			
			\item (C.4) $b_n \rightarrow 0$ as $n \rightarrow \infty$.
			
			\item (C.5) $\lambda_n \rightarrow 0$ and $\sqrt{p_n/n}/\lambda_n \rightarrow 0$.
			
			\item (C.6) There exist constants $C_1$ and $C_2$ such that when $\beta_1,\beta_2>C_1\lambda_n$, $| p\prime\prime_{\lambda_n}(\beta_1)-p\prime\prime_{\lambda_n}(\beta_2)| \leq C_2|\beta_1-\beta_2|$.
			
			\item (C.7) For Fisher information matrix $I_n(\beta_n)=E[\{\frac{\partial Q_n(\beta_n)}{\partial \beta_n}\}\{\frac{\partial Q_n(\beta_n)}{\partial \beta_n}\}^T]$, there exist constants $C_3$ and $C_4$ satisfying $0<C_3<\lambda_{min}\{I_n(\beta_n)\} \leq \lambda_{max}\{I_n(\beta_n)\}<C_4<\infty$  where $\lambda_{min}\{I_n(\beta_n)\}$ and $\lambda_{max}\{I_n(\beta_n)\}$ are minimal and maximal eigenvalues of $I_n(\beta_n)$ respectively.
			
			And for $j,k=1,2,...,p_n$,
			\[
			E_{\beta_n}\Big\{\frac{\partial Q_n(\beta_n)}{\partial \beta_{nj}}{\frac{\partial Q_n(\beta_n)}{\partial \beta_{nk}}}\Big\}^2 < C_3 < \infty
			\]
			
			and 
			\[
			E_{\beta_n}\Big\{\frac{\partial Q_n^2(\beta_n)}{\partial \beta_{nj}\partial \beta_{nk}}\Big\}^2 \leq C_4 < \infty
			\]
			
			\item (C.8) There exists a large enough open subset $w_n$ of $\Omega_n \in R^{p_n}$ which contains the true parameter point $\beta_n$, such that for almost all $V_{ni}$ the density admits all third derivatives $\partial Q_n(V_{ni},\beta_n)/\partial\beta_{nj}\partial\beta_{nk}\partial\beta_{nl}$ for all $\beta_n \in \omega_n$. Furthermore, there are functions $M_{njkl}$ such that 
			\[
			|\frac{\partial Q_n(V_{ni},\beta_n)}{\partial\beta_{nj}\partial\beta_{nk}\partial\beta_{nl}}| \leq M_{njkl}(V_{ni})
			\]
			for all $\beta_n \in \omega_n$, and 
			\[
			E_{\beta_n}\{M^2_{njkl}(V_{ni})\}<C_5<\infty
			\]
			
			\item (C.9) Let $\beta_{01},\beta_{02},...,\beta_{0k_n}$ be nonzero and $\beta_{0k_{n+1}},\beta_{0k_{n+2}},...,\beta_{0p_{n}}$ be zero. Then we have that  $\beta_{01},\beta_{02},...,\beta_{0k_n}$ satisfy
			\[
			\min\limits_{1 \leq j \leq k_n} |\beta_{0j}|/\lambda_n \rightarrow \infty \qquad \text{as} \quad n \rightarrow \infty
			\]
			
		\end{itemize}
		
		Condition (C.1) is imposed in logistic regression setting with diverging dimensionality as in (He and Shao, 2000). Condition (C.2) ensures sparse solution by making $p_{\lambda_n}(\beta_n)$ singular at the origin. Condition (C.3) guarantees unbiasedness property of large parameters, and condition (C.4) ensures $p_{\lambda_n}(\beta_n)$ does not have much more influence than $l_n(\beta_n)$ on SCAD estimator. Condition (C.5), (C.8) and (C.9) is used in the proof of Oracle property, and condition (C.6) is a smoothness condition on $p_{\lambda_n}(\beta_n)$. Condition (C.7) assumes $I_n(\beta_n)$ to be positive definite with uniformly bounded eigenvalues.

		\subsection{Proof of Theorem 3.1}
		\begin{proof}
			The proof essentially follows \cite{Fan2001}, although in this case we prove existence and consistency of the SCAD estimator via minimization of the objective function. 
			
			Let $\alpha_n=\sqrt{p_n/n}$. It's enough to show that for any given $\epsilon>0$, there exists a large constant $C$ such that
			\begin{center}
				$Pr\{\underset{\Vert{u}\Vert=C}{\text{inf}} Q_n(\beta_0+\alpha_n{u})>Q_n(\beta_0)\}\geq 1-\epsilon$
			\end{center}
			
			This implies that with probability tending to 1 there is a local minimum in the ball $\{\beta_0+\alpha_n{u}:\Vert{u}\Vert \leq C\}$ such that $\Vert\hat{\beta}_n-\beta_0\Vert=O_p(\sqrt{p_n/n})$.
			
			Denote $D_n({u})=Q_n(\beta_0+\alpha_n{u})-Q_n(\beta_0)$. Since $p_{\lambda_n}(0)=0$, we have 
			\[
			\begin{split}
			D_n({u}) = 
			L_n(\beta_0+\alpha_n{u})-L_n(\beta_0) + \Sigma^{p_n}_{j=1}[p_{\lambda_n}(|\beta_{0j}+\alpha_n{u}_j|)-p_{\lambda_n}(|\beta_{0j}|)] 
			\\
			\geq 
			L_n(\beta_0+\alpha_n{u})-L_n(\beta_0) + \Sigma^{k_n}_{j=1}[|p_{\lambda_n}(\beta_{0j}+\alpha_n{u}_j|)-p_{\lambda_n}(|\beta_{0j}|)] \\
			\stackrel{\Delta}{=} (I)  + (II)
			\end{split}
			\]
			Using Taylor's expansion, we have
			\[
			\begin{split}
			(I)=\alpha_n\triangledown^TL_n(\beta_0){u}+\frac{1}{2}{u^T}\triangledown^2L_n(\beta_0){u}\alpha_n^2+\frac{1}{6}\triangledown^T\{{u^T}\triangledown^2L_n(\beta_n^*){u}\}{u}\alpha_n^3 \\
			\stackrel{\Delta}{=} I_1 + I_2 + I_3
			\end{split}
			\]
			
			where $\beta_n^*$ lies between $\beta_0$ and $\beta_0+\alpha_n{u}$, and
			\[
			\begin{split}
			(II)=\Sigma^{k_n}_{j=1}[\alpha_np\prime_{\lambda_n}(|\beta_{0j}|)\text{sgn}(\beta_{0j})u_j+\alpha_n^2p\prime\prime_{\lambda_n}(|\beta_{0j}|)u^2_j\{1+o(1)\}]
			\\
			\stackrel{\Delta}{=} I_4 + I_5
			\end{split}
			\]
			
			According to Condition (C.3) and (C.7), 
			\[
			\begin{split}
			|I_1|=|\alpha_n\triangledown^TL_n(\beta_0){u}| \leq \alpha_n\Vert\triangledown^TL_n(\beta_0)\Vert\Vert{u}\Vert \\
			=O_p(\alpha_n/\sqrt{n})\Vert{u}\Vert=O_p(\alpha_n^2/n)\Vert{u}\Vert
			\end{split}
			\]
			
			Now we consider $I_2$. Using Chebyshev's inequality, for any $\epsilon$, we have
			\[
			\begin{split}
			P(\Vert\frac{1}{n}\triangledown^2L_n(\beta_0)-I_n(\beta_0)\Vert \geq \frac{\epsilon}{p_n}) \\
			\leq \frac{p_n^2}{n^2\epsilon^2}E\sum\limits^{p_n}_{i,j=1}\{\frac{\partial L_n(\beta_0)}{\partial\beta_{i}\beta_j}-E\frac{\partial L_n(\beta_0)}{\partial \beta_{i}\beta_j}\}^2 \\
			=\frac{p_n^4}{n}=o(1)
			\end{split}
			\]
			
			which results in $\Vert\frac{1}{n}\triangledown^2L_n(\beta_0)-I_n(\beta_0)\Vert=o_p(\frac{1}{p_n})$. Thus we have 
			
			\[
			\begin{split}
			I_2=\frac{1}{2}{u}^T[\frac{1}{n}\{\triangledown^2L_n(\beta_0)-E\triangledown^2L_n(\beta_0)\}]{u}n\alpha_n^2+\frac{1}{2}{u}^TI_n(\beta_0){u}n\alpha_n^2 \\
			= \frac{n\alpha_n^2}{2}{u}^TI_n(\beta_0){u}+o_p(1)n\alpha_n^2\Vert{u}^2\Vert
			\end{split}
			\]
			
			Based on Condition (C.8) and Cauchy-Schwarz inequality, we have
			\[
			\begin{split}
			|I_3|=|\frac{1}{6}\sum\limits^{p_n}_{i,j,k=1}\frac{\partial L_n(\beta_n^*)}{\partial \beta_{ni}\partial \beta_{nj}\partial \beta_{nk}}u_iu_ju_k\alpha_n^3| \\
			\leq \frac{1}{6}\sum\limits^{n}_{l=1}\{\sum\limits^{p_n}_{i,j,k=1}M^2_{nijk}(V_{nl})\}^{1/2}\Vert{u}\Vert^3\alpha_n^3
			\end{split}
			\]
			
			Since $p_n^4/n \rightarrow 0$ and $p_n^2\alpha_n \rightarrow 0$ as $n \rightarrow \infty$, we have 
			\[
			\begin{split}
			\frac{1}{6}\sum\limits^{n}_{l=1}\{\sum\limits^{p_n}_{i,j,k=1}M^2_{nijk}(V_{nl})\}^{1/2}\Vert{u}\Vert^3\alpha_n^3 \\
			= O_p(p^{3/2}_n\alpha_n)n\alpha_n^2\Vert{u}\Vert^2=o_p(n\alpha_n^2)\Vert{u}\Vert^2
			\end{split}
			\]
			
			Thus $I_3=o_p(n\alpha_n^2)\Vert{u}\Vert^2$.
			
			In addition, 
			\[
			\begin{split}
			|I_4| = \Sigma^{k_n}_{j=1}|\alpha_np\prime_{\lambda_n}(|\beta_{0j}|)\text{sgn}(\beta_{0j})u_j| \\
			\leq \sqrt{k_n}n\alpha_n\alpha_n\Vert{u}\Vert \leq n\alpha_n^2\Vert{u}\Vert
			\end{split}
			\]
			and
			\[
			\begin{split}
			|I_5| = \Sigma^{k_n}_{j=1}\alpha_n^2p\prime\prime_{\lambda_n}(|\beta_{0j}|)u^2_j\{1+o(1)\} \\
			\leq 2\max\limits_{1\leq j\leq k_n}p\prime\prime_{\lambda_n}(|\beta_{0j}|)\alpha_n^2\Vert{u}\Vert^2
			\end{split}
			\]
			
			By choosing a sufficiently large $C$, $I_1$, $I_3$, $I_4$ and $I_5$ are all dominated by $I_2$, which is positive. This completes the proof of Theorem 3.1.
		\end{proof}

		\subsection{Proof of Lemma}
		\begin{proof}
			It is sufficient to show that with probability tending to 1 as $n \rightarrow 0$, for any $\beta_{1n}$ satisfying $\Vert\beta_{1n}-\beta_{10}\Vert=O_p(\sqrt{p_n/n})$, for some $\epsilon_n=C\sqrt{p_n/n}$ and $j=k_n+1,...,p_n$,
			
			\[
			\left\{
			\begin{array}{ll}
			
			\frac{\partial Q_n(\beta_n)}{\partial \beta_{nj}} >0
			\qquad \qquad \text{for} \qquad 0<\beta_{nj}<\epsilon_n,\\
			
			\frac{\partial Q_n(\beta_n)}{\partial \beta_{nj}} <0
			\qquad \qquad \text{for} \qquad -\epsilon_n<\beta_{nj}<0
			
			\end{array}
			\right.
			\]
			
			
			By Taylor's expansion, 
			\[
			\begin{split}
			\frac{\partial Q_n(\beta_n)}{\partial \beta_{nj}} = \frac{\partial L_n(\beta_n)}{\partial \beta_{nj}} -p\prime\lambda_n(|\beta_{nj}|)\text{sgn}(\beta_{nj}) \\
			= \frac{\partial L_n(\beta_0)}{\partial \beta_{nj}} + \sum\limits^{p_n}_{l=1}\frac{\partial^2L_n(\beta_0)}{\partial\beta_{nj}\partial\beta_{nl}}(\beta_{nl}-\beta_{0l}) \\
			+ \sum\limits^{p_n}_{l,k=1}\frac{\partial^3L_n(\beta_n^*)}{\partial\beta_{nj}\partial\beta_{nl}\partial\beta_{nk}}(\beta_{nl}-\beta_{0l})(\beta_{nk}-\beta_{0k}) \\
			- p\prime\lambda_n(|\beta_{nj}|)\text{sgn}(\beta_{nj}) \\
			\stackrel{\Delta}{=} I_1 + I_2 + I_3 + I_4
			\end{split}
			\]
			
			where $\beta_n^*$ lies between $\beta_n$ and $\beta_0$. 
			
			For $I_1$, we have 
			\[
			I_1=O_p(\sqrt{n})=O_p(\sqrt{np_n})
			\]
			
			For $I_2$, we have
			\[
			\begin{split}
			I_2= \sum\limits^{p_n}_{l=1}\{\frac{\partial^2L_n(\beta_0)}{\partial\beta_{nj}\partial\beta_{nl}}-
			E\frac{\partial^2L_n(\beta_0)}{\partial\beta_{nj}\partial\beta_{nl}}\}
			(\beta_{nl}-\beta_{0l}) \\
			+ \sum\limits^{p_n}_{l=1}\frac{\partial^2L_n(\beta_0)}{\partial\beta_{nj}\partial\beta_{nl}}(\beta_{nl}-\beta_{0l}) \\
			\stackrel{\Delta}{=} K_1+K_2
			\end{split}
			\]
			
			Using Cauchy-Schwarz inequality and $\Vert\beta_{1n}-\beta_{10}\Vert=O_p(\sqrt{p_n/n})$, we have
			\[
			\begin{split}
			|K_2|=|n\sum\limits^{p_n}_{l=1}I_n(\beta_0)(j,l)(\beta_{nl}-\beta_{0l})| \\
			\leq nO_p(\sqrt{\frac{p_n}{n}})\{\sum\limits^{p_n}_{l=1}I_n^2(\beta_0)(j,l)\}^{1/2}
			\end{split}
			\]
			
			Since the eigenvalues of the Fisher information matrix are bounded according to Condition (C.7), we have
			\[
			\sum\limits^{p_n}_{l=1}I_n^2(\beta_0)(j,l)=O(1)
			\]
			
			Thus 
			\[
			K_2=O_p(\sqrt{np_n})
			\]
			
			For $K_1$, by Cauchy-Schwarz inequality we have
			\[
			|K_1| \leq \Vert\beta_{nl}-\beta_{0l}\Vert
			\Bigg[\sum\limits^{p_n}_{l=1}\Big\{\frac{\partial^2L_n(\beta_0)}{\partial\beta_{nj}\partial\beta_{nl}}-
			E\frac{\partial^2L_n(\beta_0)}{\partial\beta_{nj}\partial\beta_{nl}}\Big\}^2\Bigg]^{1/2}
			\]
			
			From Condition (C.7), we have
			\[
			\Bigg[\sum\limits^{p_n}_{l=1}\Big\{\frac{\partial^2L_n(\beta_0)}{\partial\beta_{nj}\partial\beta_{nl}}-
			E\frac{\partial^2L_n(\beta_0)}{\partial\beta_{nj}\partial\beta_{nl}}\Big\}^2\Bigg]^{1/2} = O_p(\sqrt{np_n})
			\]
			
			Thus $K=O_p(\sqrt{np_n})$, and $I_2=O_p(\sqrt{np_n})$.
			
			For $I_3$, we can write it as:
			\[
			\begin{split}
			I_3=\sum\limits^{p_n}_{l,k=1}\Big\{
			\frac{\partial^3L_n(\beta_n^*)}{\partial\beta_{nj}\partial\beta_{nl}\partial\beta_{nk}} - 
			E\frac{\partial^3L_n(\beta_n^*)}{\partial\beta_{nj}\partial\beta_{nl}\partial\beta_{nk}}
			\Big\}
			(\beta_{nl}-\beta_{0l})(\beta_{nk}-\beta_{0k}) \\
			+ \sum\limits^{p_n}_{l,k=1}
			E\frac{\partial^3L_n(\beta_n^*)}{\partial\beta_{nj}\partial\beta_{nl}\partial\beta_{nk}}
			(\beta_{nl}-\beta_{0l})(\beta_{nk}-\beta_{0k}) \\
			\stackrel{\Delta}{=} K_3+K_4
			\end{split}
			\]
			
			For $K_3$, by Cauchy-Schwarz inequality, 
			\[
			K_3^2 \leq \sum\limits^{p_n}_{l,k=1}\Big\{
			\frac{\partial^3L_n(\beta_n^*)}{\partial\beta_{nj}\partial\beta_{nl}\partial\beta_{nk}} - 
			E\frac{\partial^3L_n(\beta_n^*)}{\partial\beta_{nj}\partial\beta_{nl}\partial\beta_{nk}}
			\Big\}
			\Vert\beta_{n}-\beta_{0}\Vert^4
			\]
			
			Under Condition (C.8) and (C.9), we have
			\[
			K_3=O_p\Big\{\big(np_n^2\frac{p_n^2}{n^2}\big)^{1/2}\Big\}=o_p(\sqrt{np_n})
			\]
			
			For $K_4$, by Condition (C.8),
			\[
			|K_4| \leq C_5^{1/2}np_n\Vert\beta_n-\beta_0\Vert^2=O_p(p_n^2)=o_p(\sqrt{np_n})
			\]
			
			From above analysis, we have
			\[
			I_1+I_2+I_3=O_p(\sqrt{np_n})
			\]
			
			Since$\sqrt{p_n/n}/\lambda_n \rightarrow 0$ and liminf$_{n \rightarrow \infty}$liminf$_{\beta_n \rightarrow 0^+}p\prime_{\lambda_n}(|\beta_{n}|)/\lambda_n > 0$, from 
			\[
			\frac{\partial Q_n(\beta_n)}{\partial\beta_{nj}}=n\lambda_n\Big\{-\frac{p_{\lambda_n}^\prime(|\beta_{nj}|)}{\lambda_n}\text{sgn}(\beta_{nj})+O_p\big(\sqrt{\frac{p_n}{n}}/\lambda_n\big)\Big\}
			\]
			
			It is easy to see that the sign of $\beta_{nj}$ completely determines the sign of $\partial Q_n(\beta_n)/\partial\beta_{nj}$. This completes the proof. 
		\end{proof}

		\subsection{Proof of Theorem 3.2}
		\begin{proof}
			As shown in Theorem 3.1, there is a root-($n/p_n$)-consistent local minimizer $\hat\beta_n$ of $Q_n(\beta_n)$. By Lemma 3.2, part (1) holds. We only need to prove part (2), the asymptotic normality of the penalized estimator $\hat\beta_{1n}$. (Fan and Peng, 2004) showed that 
			\[
			(I_{n}(\beta_{10})+\Sigma_{\lambda_n})(\hat{\beta}_{1n}-\beta_{10})+{b_n} = \frac{1}{n}\triangledown L_n(\beta_{10})+o_p(n^{-1/2})
			\]
			
			Based on this result, we focus on its asymptotic distribution towards standard normal distribution. It is easy to see that
			
			\[
			\begin{split}
			\sqrt{n}\alpha^T_nI_{n}^{-1/2}(\beta_{10})(I_{n}(\beta_{10})+\Sigma_{\lambda_n})[(\hat{\beta}_{1n}-\beta_{10})+(I_{n}(\beta_{10})+\Sigma_{\lambda_n})^{-1}{b_n}] \\
			=\frac{1}{\sqrt{n}}\alpha_n^TI_{n}^{-1/2}(\beta_{10})\triangledown L_n(\beta_{10})+o_p(\alpha_n^TI_{n}^{-1/2}(\beta_{10}))
			\end{split}
			\]
			
			Given the conditions in Theorem 3.3, the last term is equivalent to $o_p(1)$. Let
			\[
			Y_{in}=\frac{1}{\sqrt{n}}\alpha_n^TI_{n}^{-1/2}(\beta_{10})\triangledown L_{ni}(\beta_{10}), i=1,2,...,n
			\]
			
			We consider if $Y_{ni}$ meets the conditions of Lindeberg-Feller central limit theorem. It follows that for any $\epsilon$,
			\begin{equation*}
				\begin{split}
					\sum\limits^{n}_{i=1}E\Vert Y_{in} \Vert^2 1 \{\Vert Y_{in} \Vert>\epsilon\} = nE\Vert Y_{1n} \Vert^2 1 \{\Vert Y_{in} \Vert>\epsilon\} \\
					\leq n\{E\Vert Y_{1n} \Vert^4\}^{1/2}\{P(\Vert Y_{1n} \Vert>\epsilon)\}^{1/2}
				\end{split}    
			\end{equation*}

			By Condition (C.7) and since $\alpha_n$ is an arbitrary $k_n$\textup{x}1 vector such that $||\alpha_n||=1$, we have
			\[
			P(\Vert Y_{1n} \Vert>\epsilon)) \leq \frac{E \Vert \alpha_n^TI_{n}^{-1/2}(\beta_{10}) \triangledown L_{1n}(\beta_{10}) \Vert^2}{n\epsilon}=O(n^{-1})
			\]
			
			and
			\[
			\begin{split}
			E\Vert Y_{1n} \Vert^4=\frac{1}{n^2}E\Vert \alpha_n^TI_{n}^{-1/2}(\beta_{10}) \triangledown L_{1n}(\beta_{10}) \Vert^4 \\
			\leq \frac{1}{n^2}\lambda_{max}(I_n(\beta_{10}))E\Vert \triangledown^T L_{1n}(\beta_{10}) \triangledown L_{1n}(\beta_{10}) \Vert^2 \\
			=O(\frac{p_n^2}{n^2})
			\end{split}
			\]
			
			Thus we have
			\[
			\sum\limits^{n}_{i=1}E\Vert Y_{in} \Vert^2 1 \{\Vert Y_{in} \Vert>\epsilon\}=O(n\frac{p_n}{n}\frac{1}{\sqrt{n}})=o_p(1)
			\]
			
			On the other hand, we have
			\[
			\sum\limits^{n}_{i=1}\text{cov}(Y_{in})=n\text{cov}(Y_{1n})=\text{cov}\{\alpha_n^TI_{n}^{-1/2}(\beta_{10}) \triangledown L_{1n}(\beta_{10})\} \rightarrow 1
			\]
			
			Thus $1/\sqrt{n}\alpha_n^TI_{n}^{-1/2}(\beta_{10}) \triangledown L_{1n}(\beta_{10})$ has an asymptotic standard normal distribution. This completes the proof.
		\end{proof}

		\subsection{Proof of Theorem 3.3}
		\begin{proof}
			We refer to the proof given by (Wang et. al, 2016), in which quadratic approximation of the loss function is used, so that
			\[
			Q_n(\beta_n)=\frac{1}{2}(\beta_n-\tilde{\beta_n})^T\hat{\Omega}(\beta_n-\tilde{\beta_n})+\sum\limits^{p_n}_{j=1}p_{\lambda_n}(|\beta_{nj}|)
			\]
			where $\hat{\Omega}$ is an estimate of $\Sigma^{-1}$. The asymptotic covariance matrix $\Sigma$ and its inverse matrix $\Omega$ are further decomposed into the following block matrix forms respectively according to the sparsity property in Lemma 3.2:
			\[
			\left[
			\begin{array}{cc}
			\Sigma_{11} & \Sigma_{12} \\
			\Sigma_{21} & \Sigma_{22}
			\end{array}
			\right], \left[
			\begin{array}{cc}
			\Omega_{11} & \Omega_{12} \\
			\Omega_{21} & \Omega_{22}
			\end{array}
			\right]
			\]
			
			It can be verified that
			\[
			\begin{split}
			I_n=\Omega=\left[
			\begin{array}{cc}
			\Omega_{11} & \Omega_{12} \\
			\Omega_{21} & \Omega_{22}
			\end{array}
			\right] \\
			\left[
			\begin{array}{cc}
			\Sigma_{(11)}^{-1} & -\Sigma_{(11)}^{-1}\Sigma_{12}\Sigma_{22}^{-1} \\
			=-\Sigma_{22}^{-1}\Sigma_{21}\Sigma_{(11)}^{-1}\Sigma_{22}^{-1} & \Sigma_{22}^{-1}\Sigma_{21}\Sigma_{(11)}^{-1}\Sigma_{12}\Sigma_{22}^{-1}
			\end{array}
			\right]
			\end{split}
			\]
			
			where $\Sigma_{(11)}^{-1}=\Sigma_{11}-\Sigma_{12}\Sigma_{22}^{-1}\Sigma_{21}$. Similarly, $\hat\Omega$ can also be partitioned as 
			\[
			\left[
			\begin{array}{cc}
			\hat\Omega_{11} & \hat\Omega_{12} \\
			\hat\Omega_{21} & \hat\Omega_{22}
			\end{array}
			\right]
			\]

			The existence of global minimum indicates $Pr(Q_n(\hat{\beta}_n) \leq$inf$_{\beta_n}Q_n(\beta_n)) \rightarrow 1$, which is proved in (Wang et. al, 2016) through $\hat{\beta}_{1n}=\tilde{\beta}_{1n}+\hat{\Omega}^{-1}_{11}\hat{\Omega}_{12}\tilde{\beta}_{2n}$ and we'll not further discuss here. It should be noted that a stronger condition for ensuring global minimum is obtained is discussed in (Breheny and Huang, 2011), in which the objective function $Q_n(\beta_n)$ is convex with respect to $\beta_n$ despite of the nonconvex penalty component provided that $c_*(\beta_n)>1/(\gamma-1)$, where $c_*(\beta_n)$ is the minimum eigenvalue of $n^{-1}X^TWX$, $W$ is a diagonal matrix of weights with elements $w_i=\pi_i(1-\pi_i)$, $\pi$ is the predicted probability based on most recent coordinate update, and $\gamma$ is a hyperparameter in the penalty function.
		\end{proof}
		
	\end{appendix}
	
	\section*{Acknowledgements}
	This research was partly supported by Academic Research Funding, HUB-NUS Funding and IDS Funding at the National University of Singapore.
	%
	%
	
	
	
	\bibliographystyle{imsart-nameyear} 
	\bibliography{ref} 
	
	
\end{document}